\documentclass[11pt,a4paper]{article}

\usepackage[margin=1in]{geometry}
\usepackage[T1]{fontenc}
\usepackage[utf8]{inputenc} 
\usepackage{lmodern}        
\usepackage{microtype}

\usepackage{amsmath,amssymb}
\usepackage{booktabs}
\usepackage{siunitx}

\usepackage{graphicx}
\usepackage[hyphens]{url}
\usepackage[hidelinks]{hyperref}

\usepackage{natbib}                 
\title{JSPLIT: A Taxonomy-based Solution for Prompt Bloating in Model Context Protocol}

\author{%
  Emanuele Antonioni\textsuperscript{1} \and
  Stefan Markovi\'c\textsuperscript{1} \and
  Anirudha Shankar\textsuperscript{1} \and
  Jaime Bernardo\textsuperscript{1} \and
  Lovro Markovic\textsuperscript{1} \and
  Silvia Pareti\textsuperscript{2} \and
  Benedetto Proietti\textsuperscript{1}
}

\date{} 

\newcommand{\affil}[2]{\par\noindent\textsuperscript{#1}\,#2\par}
\newcommand{\emails}[1]{\par\noindent\textbf{Email:} \texttt{#1}\par}

\begin{document}
\maketitle

\affil{1}{Janea Systems}
\affil{2}{BigFilter.ai}
\affil{}{}
\emails{emanuele@janeasystems.com; stefan@janeasystems.com; anirudha@janeasystems.com; jaime@janeasystems.com; lovro@janeasystems.com; silvia.pareti@bigfilter.ai; \\ benedetto.proietti@janeasystems.com}

\begin{abstract}
AI systems are continually evolving and advancing, and user expectations are concurrently increasing, with a growing demand for interactions that go beyond simple text-based interaction with Large Language Models (LLMs). Today’s applications often require LLMs to interact with external tools, marking a shift toward more complex agentic systems. To support this, standards such as the Model Context Protocol (MCP) have emerged, enabling agents to access tools by including a specification of the capabilities of each tool within the prompt. Although this approach expands what agents can do, it also introduces a growing problem: prompt bloating. As the number of tools increases, the prompts become longer, leading to high prompt token costs, increased latency, and reduced task success resulting from the selection of tools irrelevant to the prompt.
To address this issue, we introduce JSPLIT, a taxonomy-driven framework designed to help agents manage prompt size more effectively when using large sets of MCP tools. JSPLIT organizes the tools into a hierarchical taxonomy and uses the user’s prompt to identify and include only the most relevant tools, based on both the query and the taxonomy structure.
In this paper, we describe the design of the taxonomy, the tool selection algorithm, and the dataset used to evaluate JSPLIT. Our results show that JSPLIT significantly reduces prompt size without significantly compromising the agent’s ability to respond effectively. As the number of available tools for the agent grows substantially, JSPLIT even improves the tool selection accuracy of the agent, effectively reducing costs while simultaneously improving task success in high-complexity agent environments.
\end{abstract}

\section{Introduction}
\label{sec:introduction}
During the past year, artificial intelligence has undergone a remarkable shift. What was once defined largely by conversational tools like Large Language Models (LLMs) to respond to user questions in a one-to-one exchange is now evolving into something more complex and dynamic: AI agents.
AI agents are fundamentally different from normal conversational AIs \cite{zhu2025evolutionary}: they're built to take action on their own using a variety of tools. They can plug into APIs, search databases, fill out spreadsheets, work within CRM systems, move through cloud environments, and much more, without waiting for a human to guide every step.  AI agents do not only answer questions, but handle entire autonomous workflows, making decisions and adapting in real time to environmental state changes \cite{yao2023react}.
\begin{figure}[t]
\centering
\includegraphics[width=0.4\columnwidth]{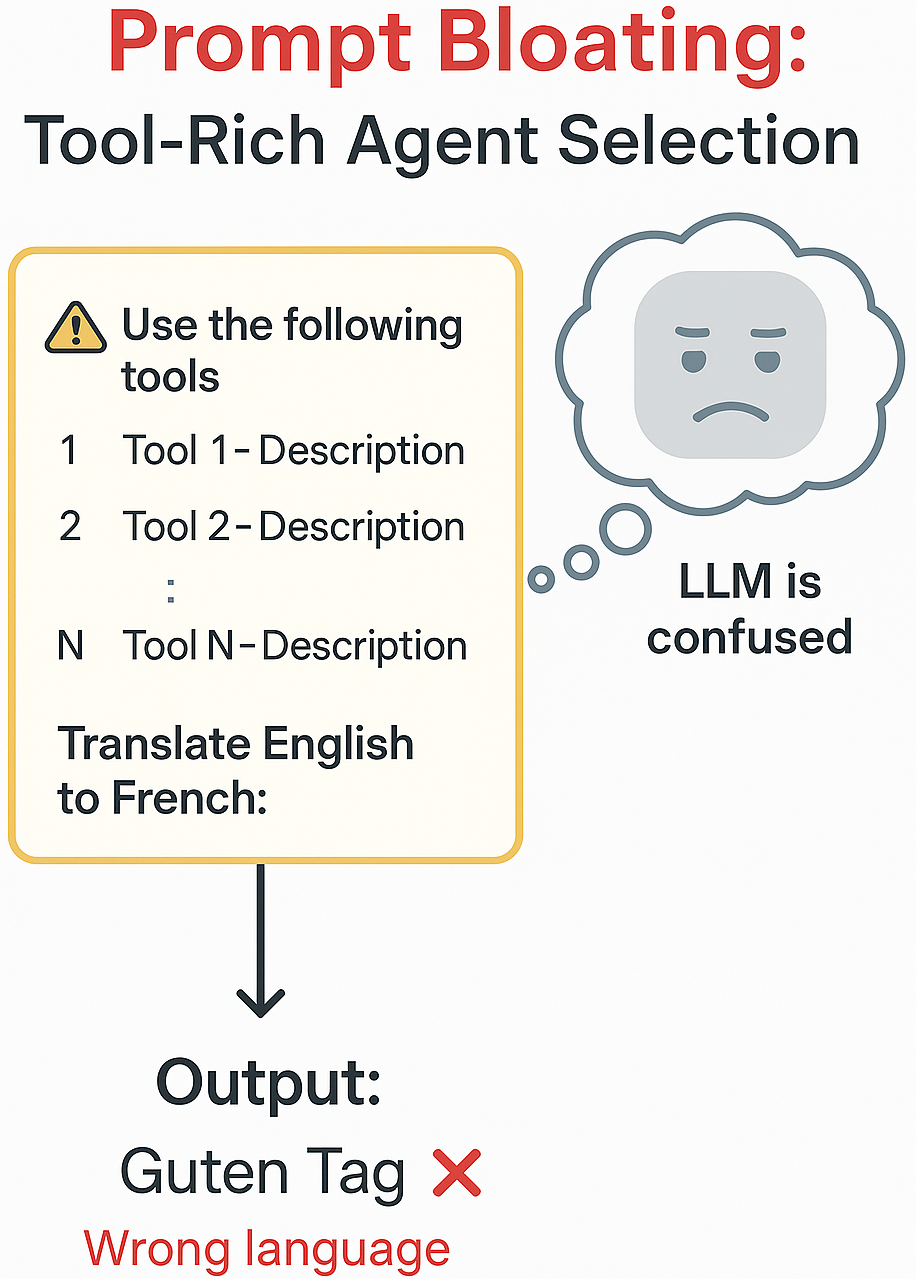} 
\caption{An example of prompt bloating caused by a massive amount of tools descriptions injected into the agent's context}
\label{fig:prompt_bloating}
\end{figure}
This has been framed by several enterprise investigations. According to the PwC 2025 report \cite{pwc2025}, three out of four business leaders believe that AI agents could bring about a disruptive transformation, stating that they can even overcome the impact of smartphones and the internet. In the same research, more than 80\% companies plan to increase their investments in agent-based systems over the next two years.
A similar research carried out by Capgemini in July 2025 \cite{capgemini2025} supports this momentum. They found that 14\% of organizations around the world have already deployed AI agents in live environments, while another 23\% are piloting them.
What is emerging is a new model for how humans and machines work together. In this new paradigm, the AI agent can make autonomous decision even without the full control of the human operator. These systems can detect changes in context, decide what actions to take, and even collaborate with other agents when necessary. 
Following the increasing interest of the entire community around the integration of AI agents, different dedicated tools and technologies arose in the latest year. Between agentic AI technologies Model Context Protocol (MCP) \cite{MCP} usage has spread significantly receiving support from big partners such as OpenAI, Microsoft and Google. The Model Context Protocol serves as a universal interface for connecting LLMs with external data sources and tools. It eliminates the need to develop custom integrations for each model-tool combination, instead offering a standardized and secure connection based on JSON-RPC 2.0. The protocol defines a client-server architecture: the AI agent (MCP client) connects to one or more MCP servers that expose functionality via contextual metadata and predefined functions.
This new instrument brings its own set of challenges. One structural problem of this approach is the prompt bloating (Figure \ref{fig:prompt_bloating}).
Prompt bloating \cite{addad2024homeopathic, long2025eperm} is a phenomenon that emerges frequently in the practical use of AI, particularly when trying to extend the conversational context or deliver complex instructions through increasingly long textual inputs due to the context injection of tools' descriptions. While this approach makes it possible to maintain a better control on the behavior of the model, it also introduces a number of significant negative effects, both in terms of computational and token costs, and the quality of the responses generated. 
One of the main disadvantages concerns the increase in computational cost. LLM models typically have a computational complexity quadratic to the length of the prompt. This means that doubling the length of the prompt can quadruple the time and memory required to process it.
In addition to the infrastructural impact, prompt bloating also has direct economic consequences when using models accessible through paid APIs. In these cases, the cost of inference is generally calculated based on the total number of tokens processed. As a result, an unnecessarily long prompt not only increases computational requirements but also proportionally affects the actual price of the interaction.
Qualitatively, prompt bloating tends to worsen the accuracy and relevance of model responses. With an overly long prompt, the model is exposed to information that is potentially contradictory, obsolete, or simply irrelevant to the current task. This cognitive overload compromises the model's ability to focus on important details, leading to worse performance in terms of general response and tool selection accuracy. Nowadays users adopt many pragmatic solutions for handling long prompts. Among these, one of the most popular is periodic context summarization \cite{wang2024adapting}, in which portions of previous text or dialogues are automatically summarized to reduce the number of tokens. Other strategies include the use of sliding windows \cite{liimproving} to maintain only the most recent and relevant context, and semantic prioritization, which dynamically selects the most relevant information with respect to the current task. These techniques, even if effective cannot be directly applied to AI agents, because the tools description are often already short and hard to summarize, and also it is impossible to cut out parts of the description using an agnostic approach without seriously compromising the agent's chance of selecting the right tool.
To effectively mitigate the problem of prompt bloating in AI agents interfacing with a large number of MCP servers, we introduce JSPLIT, a novel system designed to optimize context selection through intelligent taxonomy-based filtering. The core idea behind JSPLIT is to organize all available MCP servers into a hierarchical taxonomy, where each taxonomy class is associated with a human-readable description that captures its functional scope.
This taxonomy is pre-integrated into the system and serves as a structural map to guide contextual pruning. When a user query is received, the large language model embedded in the agent evaluates the query against the set of class descriptions in the taxonomy. Based on semantic relevance, it selects only those taxonomy classes that are pertinent to the query. The system then filters the MCP server pool accordingly, selecting only the servers that belong to the identified classes. These selected servers—and only these—are included in the agent’s execution context.
By narrowing the context to only the most relevant subset of MCP servers, JSPLIT significantly reduces prompt size while preserving functional completeness for the task at hand. Based on this framework, in this paper we introduce the following contributions:
\begin{itemize}
    \item A set of different taxonomies to divide the MCP servers
    \item An evaluation dataset of thousands of MCP servers classified according to the taxonomies
    \item The JSPLIT system used for selecting the servers according to the user's query
    \item A test dataset that connects queries with the correct server to use for completing the task
\end{itemize}
The remainder of this paper is organized as follows. In the \textit{Related Work} section, we review related work on prompt management applied to AI agents architectures. Section \textit{Method} introduces our proposed approach, JSPLIT, with a discussion of the  JSPLIT algorithm. Section \textit{Datasets and Taxonomies} describes the different dataset we provided for evaluating the approach and the taxonomies we developed for classifying the MCP servers used in the evaluation. Section \textit{Results} presents experimental results that demonstrate the effectiveness of our method in reducing context size while maintaining agent performance. In Section \textit{Error Analysis}, we conduct an error analysis to better understand the limitations and failure cases of the approach. Finally, Section \textit{Conclusion and Future Work} summarizes our findings and outlines directions for future research.

\section{Related Work}
\label{sec:relwork}
The growing adoption of AI agents capable of interfacing with external tools and services has introduced pressing challenges around context management. As these agents are tasked with increasingly complex and diverse operations, the volume of supporting information required for effective reasoning and decision-making can quickly lead to severe prompt bloating. In response, recent research has explored strategies for mitigating context overload, particularly in the areas of task selection and scalable prompt construction within modular AI agent systems.
A key benchmark in this space is introduced in \cite{robertsneedle}, where the authors evaluate large language models’ (LLMs) ability to retrieve and reason over information spread across contexts approaching a million tokens. Their work focuses on needle threading tasks—scenarios in which models must follow long chains of related information buried within vast prompt windows. They demonstrate that retrieval performance tends to degrade as context size increases, underscoring a fundamental limitation that motivates the core problem addressed in this paper.
To address the challenge of effective tool selection, \cite{kachuee2025improving} propose enhancing retrieval through LLM-based query generation. Rather than depending exclusively on dense retrievers or embedding similarity, their approach leverages the LLM's own contextual reasoning abilities to generate more targeted queries. They explore zero-shot prompting, supervised fine-tuning, and alignment learning—ultimately showing that alignment learning provides robust performance, particularly in out-of-domain scenarios. Their findings suggest that LLMs can serve not only as language generators but also as intelligent retrievers of their own operational context. Building on this idea, our work introduces a taxonomy-based selection mechanism that further improves retrieval success rates in long prompts while also reducing token usage throughout the agent's operation.
Complementary approaches to tool selection have also been explored. \cite{gao2024confucius} present Confucius, a tool-learning framework designed to enhance LLMs’ ability to interact with complex tools in real-world tasks. Their curriculum-based training procedure—consisting of warm-up, in-category, and cross-category phases—gradually teaches tool use while refining the training data via introspective feedback (ISIF), which helps the model focus on nuanced tool behaviors.
Another related line of work is ToolkenGPT, introduced by \cite{hao2023toolkengpt}, which takes a different approach by learning new embeddings to support tool selection. Their framework augments frozen language models with a wide array of external tools by learning tool embeddings that integrate into the generation process. ToolkenGPT employs a two-stage training strategy that aligns tool usage with the language modeling objective, enabling parameter-efficient inference without modifying the base model’s weights.
Unlike these approaches, which require model fine-tuning or embedding updates, our method (JSPLIT) operates post-training without altering the LLM’s parameters. This makes our approach more modular and broadly applicable, especially in settings where model weights are inaccessible or frozen.

\section{Method}
\label{sec:method}
JSPLIT is an intelligent agent framework designed to run AI agents by combining large language model (LLM) reasoning with external tool execution through a structured and explainable control pipeline. Users provide textual queries or task descriptions, which JSPLIT resolves by orchestrating interactions between an LLM and a suite of tool servers that connect the system to the external world.
At the core of JSPLIT is the Taxonomy-MCPResolver, a module responsible for intelligently selecting the most relevant tool servers based on a semantic classification of the user’s query. This selective routing enables efficient and context-aware execution of complex tasks by ensuring that only the appropriate tools are invoked during the resolution process.
In real-world applications, JSPLIT is expected to be integrated into general-purpose AI orchestration frameworks that support tool-augmented reasoning. These frameworks facilitate iterative interactions between LLMs and external tools, allowing complex tasks to be decomposed, delegated, and refined until a satisfactory result is achieved.
\subsection{System Overview}
JSPLIT is a modular framework for AI agents that combines LLM reasoning with access to large pools of external services, known as Model Context Protocol (MCP) servers. It enables efficient and scalable tool use through intelligent server selection and iterative query resolution.
When a user query enters the system, the Taxonomy-MCPResolver determines which MCP servers are relevant. It uses a hierarchical taxonomy that classifies servers by functionality and matches the query to human-readable class descriptions. Only the servers mapped to the most semantically appropriate categories are retained for further processing.
The selected MCP servers and the query are then passed to the LLM, which attempts to resolve the task through an iterative process called the call loop. At each step, the LLM decides whether to directly answer the query based on its current context or to invoke tools on the selected MCP servers. If tools are invoked, their outputs are appended to the prompt, updating the context for the next iteration. The loop continues until the LLM produces an answer or a maximum number of iterations is reached.
The final output includes the LLM’s answer (if one was generated), the list of MCP servers used, and token usage statistics.
JSPLIT also supports a baseline mode in which the Taxonomy-MCPResolver is replaced with a Passthrough-MCPResolver. In this configuration, no filtering is applied, and all MCP servers are made available to the LLM. This allows performance comparisons between intelligent and unfiltered tool selection strategies.
In the next section we will explore more in detail the core component of the JSPLIT system: the Taxonomy-MCPResolver

\begin{figure}[t]
\centering
\includegraphics[width=0.9\columnwidth]{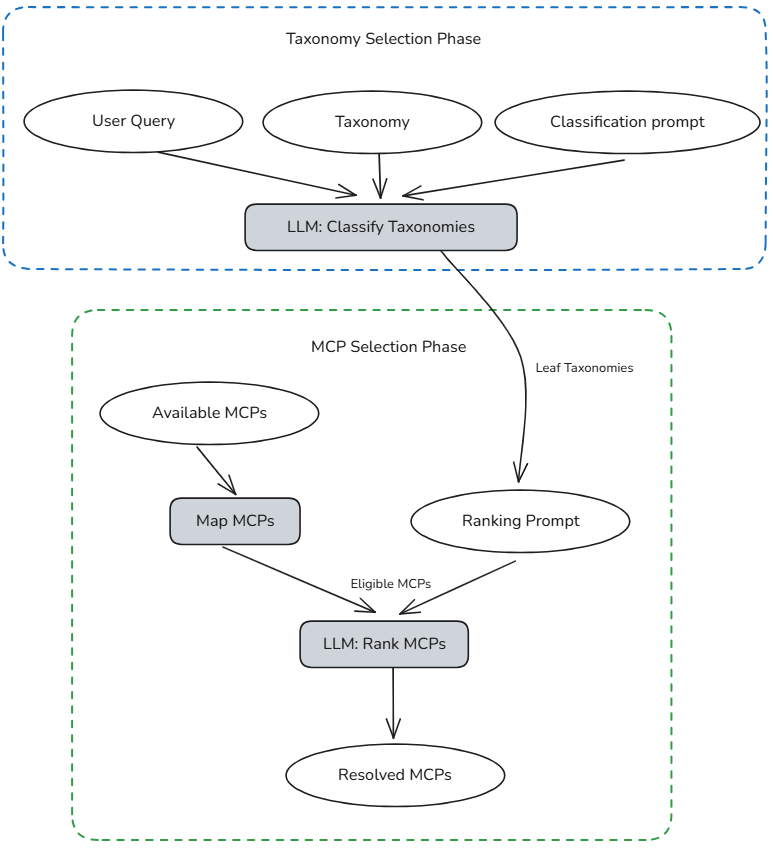} 
\caption{The overall logic of the Taxonomy-MCPResolver component}
\label{fig:taxres}
\end{figure}
\subsection{Taxonomy Resolver}

This component performs intelligent selection of Model Context Protocol (MCP) servers by combining a hierarchical taxonomy of the servers with the large language model classification.
The resolver is instantiated with a configuration file defining available MCP servers, a JSON-formatted taxonomy file, and a client interface to a large language model (LLM).
The core resolution logic follows a two-phase sequence: taxonomy classification and MCP selection.

In the taxonomy classification phase, the system first preprocesses the taxonomy by filtering it to retain only categories that have at least one associated MCP in the current agent's state, then formats the remaining structure into a hierarchical string structured as a tree. It constructs a classification prompt by inserting this formatted taxonomy into a predefined template that instructs the LLM to choose the most specific leaf-level category for the input query. The query and prompt are submitted to the LLM, whose output is parsed to extract a valid taxonomy identifier corresponding to the selected category. 

In the MCP mapping and LLM-based ranking phase, the resolver retrieves MCPs directly mapped to the identified taxonomy category via dictionary lookup; if only one MCP matches, the resolver selects it directly. If multiple MCPs are eligible, the resolver generates a ranked-list prompt that describes each candidate with a truncated summary and asks the LLM to rank the options, returning a comma-separated list of indices for the top-k MCPs. The system validates the returned indices, uses them to index into the candidate list, and assembles the selected MCPs into the final result. 
Figure \ref{fig:taxres} illustrates the overall schema of the Taxonomy Resolver

\section{Datasets and Taxonomies}
In this section, we introduce the core datasets and taxonomies that represent the backbone of the JSPLIT system. These foundational resources enable the platform to classify tools, resolve queries, and guide language model reasoning in a structured and interpretable way. The datasets provide labeled examples of both tools and queries, while the taxonomies offer a hierarchical framework for organizing and navigating the tool landscape. Together, they support the development, evaluation, and continuous improvement of JSPLIT’s resolution and orchestration capabilities.

\subsection{Taxonomies}
To support structured tool reasoning and precise query resolution, the JSPLIT system relies on a functional taxonomy—a hierarchical classification system that organizes MCP (Model Context Protocol) servers based on what they do. This taxonomy enables the language model to reason over tool capabilities, supports accurate prompt generation, and guides the system in selecting the most relevant services for a given user query. Over the course of JSPLIT’s development, two versions of the taxonomy have been created: Taxonomy v1 and a more advanced Taxonomy v2.

Taxonomy v1 was the first structured attempt to categorize the growing ecosystem of MCP servers. It was designed with simplicity and flexibility in mind, using a hierarchical structure based on primary functionality, supported by optional secondary tags for data type (e.g., text, image, structured) and provider (e.g., OpenAI, Microsoft). Each server was assigned a single primary category from a three-level hierarchy, with additional tags allowed for multi-functional tools.
The taxonomy emphasized eight broad functional areas, including search, knowledge management, data processing, simulation, and communication. It supported multi-tool tagging, allowing servers to reflect multiple capabilities without violating the core hierarchy. Taxonomy-v1 served as a preliminary foundation, enabling early experimentation with tool classification and routing, and ultimately guiding the development of the more comprehensive and structured Taxonomy-v2.

Taxonomy v2 builds directly over v1 but introduces a deeper and more structured framework. It expands the number of top-level categories to eleven, covering new domains such as developer tools, specialized industries (e.g., finance, entertainment), and multi-domain orchestration. Each subcategory is now paired with a clear definition, ensuring consistency in both manual and automated classification.
Taxonomy v2 introduces fallback subcategories for tools that don’t neatly fit into existing slots, providing a formal mechanism for handling outliers. It also improves clarity in naming and scope, and it integrates hybrid and cross-functional tools more explicitly.

\subsection{Datasets}
The JSPLIT system is developed and evaluated using two complementary datasets that reflect different aspects of its functionality: the MCP Server Dataset and the MCP Query Dataset. Each serves a distinct role—one capturing the structure and capabilities of the tool landscape, and the other modeling the types of tasks users might pose to the system.

The MCP Server Dataset is a comprehensive catalog of approximately 2,000 Model Context Protocol (MCP) servers, retrieved from Smithery, a widely used MCP registry. Each MCP represents an external tool or service that can be invoked during the JSPLIT query resolution process. Every server entry includes a name, a human-readable description of its functionality, and a machine-readable specification of the tools it provides, all formatted according to the JSON schema defined in the MCP protocol.
To support structured tool selection, we categorized all MCP servers under a multi-level taxonomy that reflects their functional domains and capabilities. This taxonomy underpins the intelligent routing performed by JSPLIT’s Taxonomy-MCPResolver, enabling the system to reason about which tools are most appropriate for a given task. Classification of the dataset was conducted in two stages: an initial subset was manually annotated by domain experts to ensure high-quality labels, and these examples were then used to guide automated expansion using an external language model (Claude Sonnet 3.7). The LLM generalized from the expert-labeled entries to assign taxonomy categories to the remaining servers, resulting in a fully labeled and operational dataset.

The MCP Query Dataset contains approximately 200 entries, each representing a realistic user query paired with a ground-truth annotation. For each query, the dataset specifies the correct MCP server that should be used to fulfill the request, along with the appropriate taxonomy category. This dual annotation provides a structured foundation for evaluating both tool selection accuracy and end-to-end task resolution within the JSPLIT pipeline.
The dataset is constructed to rigorously test the full capabilities of JSPLIT, including query interpretation, taxonomy-based reasoning, and tool invocation. By defining both the intended outcome and the taxonomy context, it supports detailed evaluation of intermediate steps as well as overall task success.
Roughly 50\% of the entries were initially generated using a language model and then manually refined to ensure they contained all necessary information for tool invocation. These queries were carefully adjusted for clarity and confirmed to succeed when evaluated in isolation (i.e., without irrelevant tools present). The remaining 50\% were automatically generated by selecting MCP tools and producing corresponding queries with the goal of maximizing taxonomy coverage. Each automatically generated query was validated end-to-end through the JSPLIT system, and only retained if the correct tool was successfully invoked in a clean execution environment.

\section{Experimental setup}
To evaluate JSPLIT’s performance, a structured experimental setup was designed to simulate an AI orchestration pipeline and test the system under various configurations. The experiments aimed to measure both the correct selection and invocation of MCP servers, and the LLM-related costs, such as token usage.

Each experiment followed a "needle in a haystack" design: for every user query in the MCP Query Dataset, the correct target MCP server (the "needle") was embedded among a set of irrelevant, randomly sampled "noise" MCP servers (the "haystack"). These noise MCPs were drawn from a larger pool of approximately 2,000 MCP servers, excluding any from the same leaf-level taxonomy category as the target. To assess scalability and robustness, the number of noise MCPs was varied incrementally, from 1 up to 1,000, allowing the system’s performance to be evaluated under increasing levels of task complexity and tool-space clutter.

For each query: (1) a list of available MCPs was constructed by mixing the target server with a randomly selected set of noise MCPs. (2) The user query and this list were passed into the JSPLIT system.(3) JSPLIT executed its full pipeline, including the Taxonomy-MCPResolver (if enabled), LLM reasoning, and tool invocation.
(4) The output was recorded, and LLM token usage and estimated costs were logged.
(5) Accuracy was computed by checking whether the target MCP was correctly invoked.

Over 95\% of outputs were tool-call executions. An output was marked correct if any tool from the target MCP server was used. For the minority of direct LLM answers ($<$5\%), a LLM-as-a-judge method was employed: a separate LLM was asked to evaluate the correctness of the answer by comparing it to the known ground truth.
This setup also enabled controlled comparison between two configurations of JSPLIT: one using the Taxonomy-MCPResolver for intelligent server filtering, and one using the Passthrough-MCPResolver, which passes all MCP servers to the LLM without filtering. This allowed for a quantitative assessment of the benefits of taxonomy-based routing in terms of both effectiveness and computational efficiency.
\begin{figure}[t]
\centering
\includegraphics[width=0.9\columnwidth]{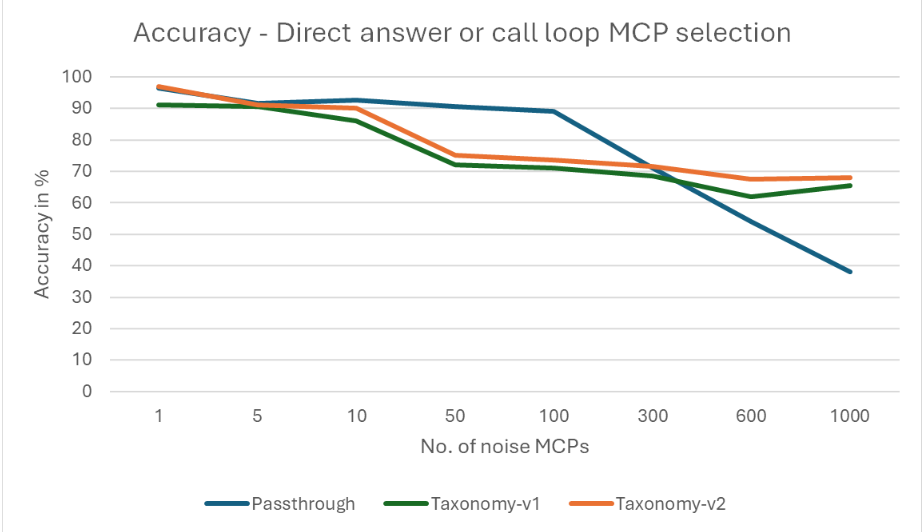} 
\caption{Accuracy of correct tool selection over the number of MCP servers. "Passthrough" represent the vanilla system that embeds all description into the context, "Taxonomy-v1" refers to the use of JSPLIT using the v1 version of the taxonomy, and "Taxonomy-v2" represents JSPLIT using the v2 taxonomy}
\label{fig:accuracy}
\end{figure}
\section{Results}
In this section, we evaluate the effectiveness of the JSPLIT system in reducing input token cost and improving tool selection accuracy when the number of MCP server connected to the agent arise. Our experiments are designed to assess both the overall performance of taxonomy-based filtering compared to baseline approaches, and the sensitivity of the system to variations in the underlying model used during the filtering phase.

\label{sec:results}

\begin{figure}[t]
\centering
\includegraphics[width=0.9\columnwidth]{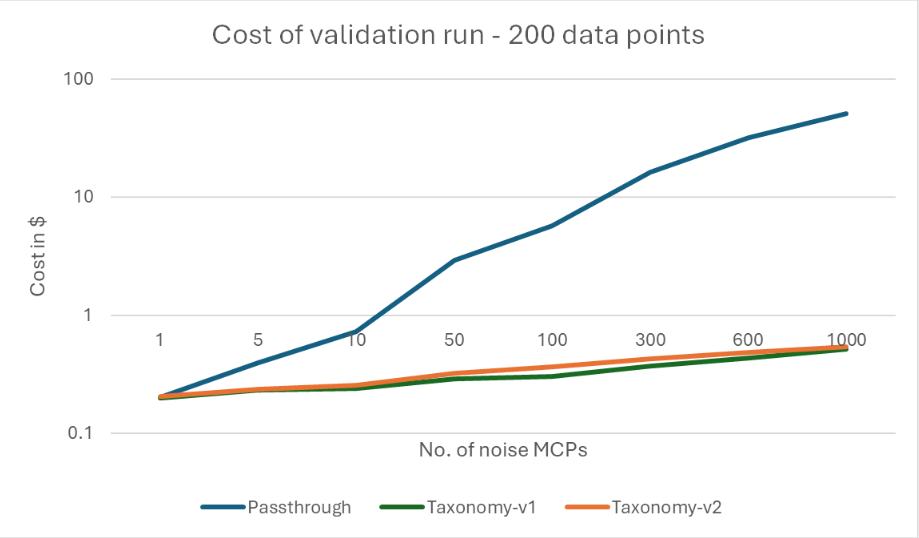} 
\caption{Input token cost for running 200 queries to the LLM over the number of MCP servers. "Passthrough" represent the vanilla system that embeds all description into the context, "Taxonomy-v1" refers to the use of JSPLIT using the v1 version of the taxonomy, and "Taxonomy-v2" represents JSPLIT using the v2 taxonomy}
\label{fig:token_cost}
\end{figure}

The first experiment evaluates three system configurations: (1) a baseline condition in which all MCP server descriptions are directly injected into the LLM context (Passthrough); (2) the JSPLIT system employing Taxonomy-v1 for tool filtering; and (3) the same system utilizing the more refined Taxonomy-v2. In all configurations, GPT-4.1-mini is used as the language model across all AI interaction steps.
As illustrated in Figure \ref{fig:accuracy}, JSPLIT demonstrates improved tool selection accuracy over the baseline when the number of MCP servers is low (fewer than five). A slight decrease in performance is observed as the number of servers increases over ten MCP servers, corresponding to a rapid expansion in the number of candidate taxonomy classes injected in the LLM prompt during the selection phase. However, as the server pool continues to grow—reaching into the hundreds—the performance of JSPLIT stabilizes, whereas the accuracy of the Passthrough approach deteriorates markedly, with the Passthrough approach culminating on less than 40\% accuracy, while JSPLIT with Taxonomy-v2 remains stable around 69\%. Taxonomy-v2 show comparable performance with Taxonomy-v1 when the noise servers are below five, but it demonstrate better performance from then on.
Figure \ref{fig:token_cost} reports the cumulative cost (in USD) of input tokens required to complete a batch of 200 queries as the number of connected MCP servers increases. The results clearly indicate that JSPLIT achieves a substantial reduction in token cost—exceeding two orders of magnitude—compared to the baseline. This demonstrates the effectiveness of JSPLIT in controlling computational and financial overhead while scaling to large tool ecosystems.

\begin{figure}[t]
\centering
\includegraphics[width=0.9\columnwidth]{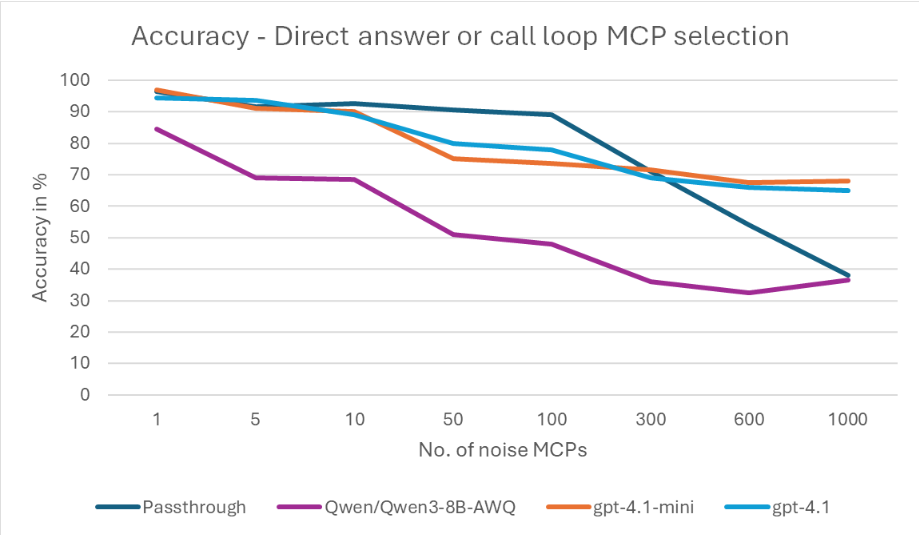} 
\caption{Accuracy of tool selection of different models used for the inner loop of the JSPLIT system. }
\label{fig:models}
\end{figure}

\begin{figure}[t]
\centering
\includegraphics[width=0.9\columnwidth]{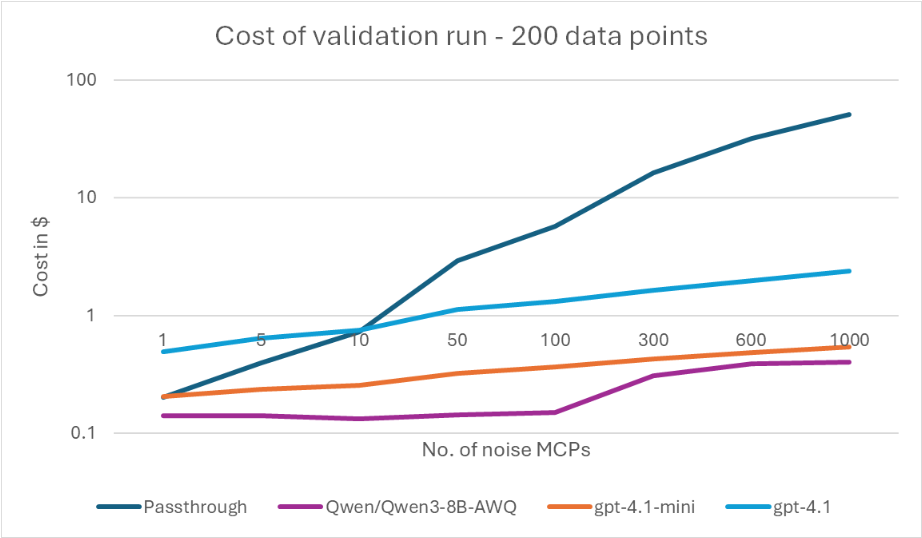} 
\caption{Input token cost for running 200 queries to the LLM over the number of MCP servers using different models for the MCP filtering step of the JSPLIT system}
\label{fig:models_token_cost}
\end{figure}

The second experiment is an ablation study designed to assess how the choice of language model within JSPLIT’s inner loop component responsible for selecting relevant taxonomy classes and MCP tools—affects overall system performance. Specifically, we compare three variants of JSPLIT, each using a different model for the classification step: a local LLM (Qwen3-8B-AWQ), a small API-based model (GPT-4.1-mini), and a larger API-based model (GPT-4.1). The final interaction step is completed using GPT-4.1-mini in all three environments.
Figure \ref{fig:models} presents the results in terms of tool selection accuracy. The findings show that both API-based models achieve comparable levels of performance, whereas the local model results in a substantial drop in accuracy. This suggests that while local models can support inference at lower cost, they may struggle to generalize effectively in complex selection tasks.
In Figure \ref{fig:models_token_cost}, we report the input token cost (in USD) associated with each variant. As expected, the local model offers the lowest cost, but its savings are offset by a significant decrease in accuracy. The small API model provides a strong balance, maintaining high accuracy while keeping costs relatively low. The use of the larger API model yields only a marginal improvement in accuracy over the smaller one, but incurs a noticeable increase in token-related expenses. These results highlight a trade-off between computational cost and model performance, suggesting that lightweight API models may offer the best compromise for many practical deployments.  

\begin{figure}[h]
\centering
\includegraphics[width=0.9\columnwidth]{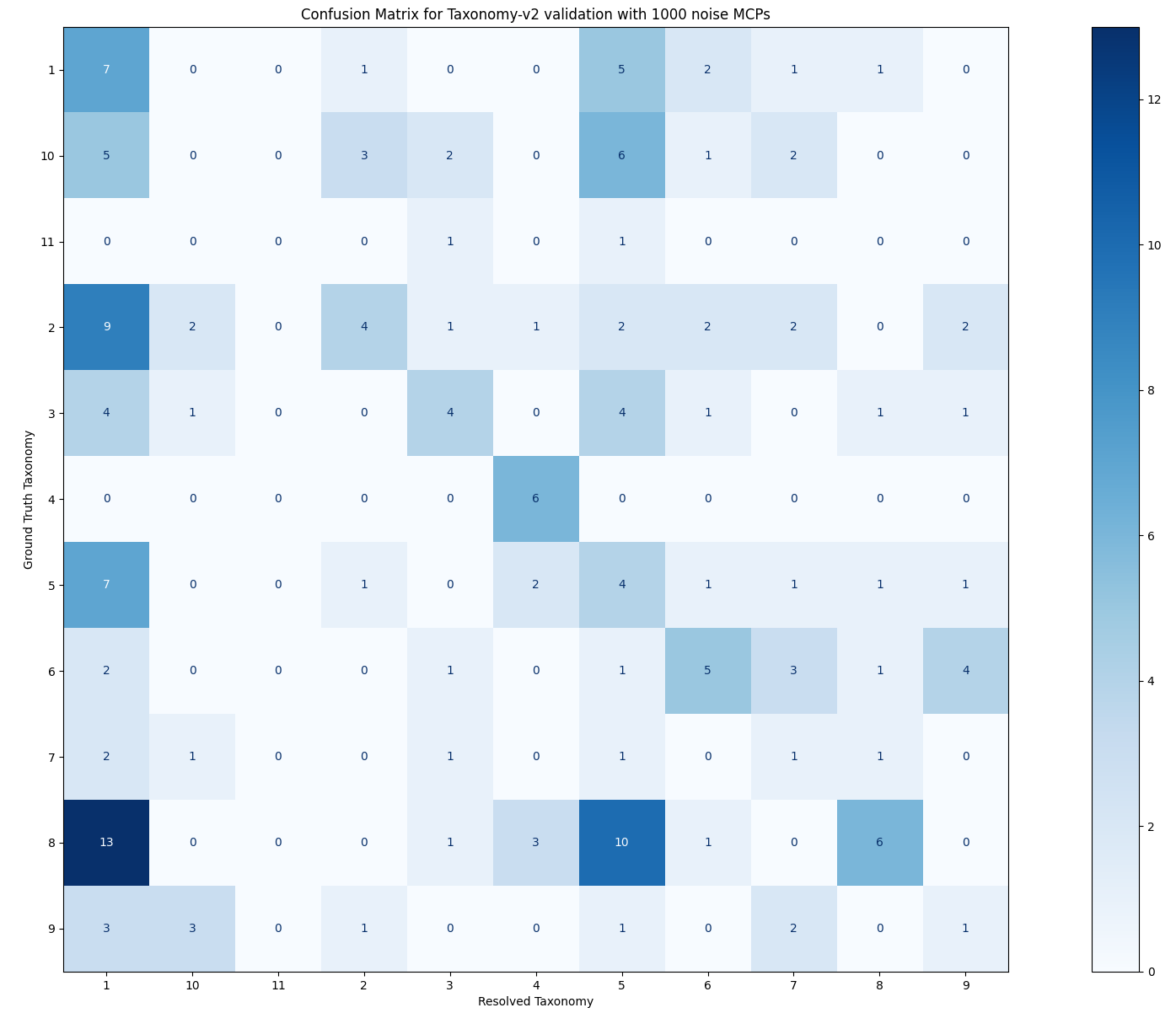} 
\caption{Confusion matrix representing classification errors between top-level classes of the taxonomy-v2}
\label{fig:error_analysis}
\end{figure}

\section{Error Analysis}
\label{sec:erran}
The confusion matrix in Figure~\ref{fig:error_analysis} provides insight into the classification behavior of the JSPLIT system when using Taxonomy v2 in the presence of 1000 noise MCPs not relevant to the target queries. Each row corresponds to the ground truth top-level taxonomy class, while each column shows the predicted top-level class. Values off the diagonal indicate misclassifications at the top category level, while non-zero diagonal entries reflect correct top-level classification with errors at deeper levels of the taxonomy.
The top classes are namely: (1)Search and information retrieval, (2)Memory and knowledge management, (3)Simulation and planning, (4)Navigation and mapping, (5)Data extraction and manipulation, (6)System and device control, (7)Communication and interaction, (8)Specialized domains, (9)Developer tools and programming, (10) Multi-domain orchestration, (11) Others.

Several classes exhibit consistent confusion with semantically similar or overlapping categories:
Memory and Knowledge Management (2) is frequently confused with Search and Information Retrieval (1) and Multi-Domain Orchestration (10). This likely stems from the overlap between tools that store and retrieve personal knowledge and those used for general search or coordination across domains.
Simulation and Planning (3) shows notable misclassification toward Data Extraction and Manipulation (5) and System and Device Control (6), possibly due to the presence of data-driven decision tools and modeling systems that share similar operational features.
These patterns highlight the challenge of distinguishing categories that involve compound or overlapping workflows, especially under weak signal conditions or when server descriptions are brief or ambiguous.

The category Search and Information Retrieval (1) appears across multiple rows in the matrix, indicating that it is often selected even when it is not the correct ground truth class. Misclassifications from categories such as Data Extraction and Manipulation (5), Specialized Domains (8), and Memory and Knowledge Management (2) frequently target this class. This suggests that descriptions of tools involving data access, content lookup, or general retrieval behavior may be semantically close to those found in search-related services. This over-selection highlights the need for more discriminative features or clearer category boundaries between search and related data access functionalities.
The Specialized Domains (8) category suffers from significant misclassification, particularly into Search and Information Retrieval (1) and System and Device Control (6). This is unsurprising, as tools in specialized domains (e.g., finance, healthcare) often reuse common backend technologies and may not be distinguishable without highly domain-specific context.

\section{Conclusion and Future Work}
\label{sec:conclusion}
In this paper, we introduced JSPLIT, a system designed to address the challenge of prompt bloating in AI agent architectures that operate over large pools of external tools. Using a taxonomy-based filtering mechanism, JSPLIT dynamically selects a subset of relevant MCP servers to include in the language model’s context, significantly reducing token usage while preserving the agent’s operational effectiveness.
Our evaluation, based on a large dataset of classified MCP servers and a set of task-linked queries, demonstrates that JSPLIT consistently reduces prompt size without significantly sacrificing task accuracy. In particular, as the number of available servers scales into the hundreds, the system shows clear advantages over baseline approaches that inject the full tool context. Notably, JSPLIT achieves better tool selection accuracy under these high-density conditions, highlighting the importance of structured pruning in this type of agent environment.
While these results are encouraging, several areas remain open for improvement. One key direction involves refining the descriptions associated with taxonomy categories, which play a central role in semantic matching between user queries and tool functions. Improving these descriptions, both in clarity and coverage, could further enhance the quality of classification and context filtering.
Additionally, we plan to explore the development of a real-time classification mechanism for onboarding new MCP servers. Currently, classification relies on a static initial annotation, which limits the system’s adaptability in dynamic environments. A live, model-assisted classification pipeline would allow JSPLIT to scale more fluidly and remain aligned with evolving tool ecosystems.
In parallel, we have begun work on a more sophisticated Taxonomy-v3, which introduces more independent categories and separates domain as its own classification dimension. While still under development and requiring further experimentation, the design of Taxonomy-v3 shows promise as a more flexible and expressive foundation for future iterations of JSPLIT.

Overall, this work demonstrates that taxonomy-guided context management is a viable and effective strategy for improving the scalability and reliability of AI agents working with large sets of tools and MCP servers. Continued improvements in adaptability and taxonomy structure are likely to further extend the system’s capabilities in real-world deployments.

\bibliography{bibliography}

\newpage
\section*{\LARGE APPENDIX A}
\section*{Taxonomy-v1}

\subsection*{Approach}
\begin{itemize}
    \item \textbf{Primary Axis:} Functionality (What the server does)
    \item \textbf{Secondary Dimensions:} Data Type, Provider
\end{itemize}

\subsection*{Handling Multi-Tool Servers}
\begin{itemize}
    \item Allow servers to be tagged with multiple categories.
    \item This deviates from ``Single Inheritance'' but enables recognition of ambiguity and allows proposing alternatives or refining requests.
\end{itemize}

\subsection*{Primary Classification Rules}
\begin{itemize}
    \item \textbf{Dominant Function:} Classify by the server's primary or most prominent capability.
    \item Each MCP server must have a single, primary classification (Level 1, 2, or 3) to preserve the search hierarchy.
    \item \textbf{Secondary Function:} Additional functionality tags are allowed.
    \item \textbf{Example:} A server primarily offering ``1.1.1 General Web Search'' but also ``5.1.1 NLP'' is classified under 1.1.1 and tagged with 5.1.1.
\end{itemize}

\subsection*{1. SEARCH AND INFORMATION RETRIEVAL}
\textbf{Definition:} Servers that locate, fetch, and deliver information from various sources.

\subsubsection*{1.1 Web Search and Discovery}
\begin{itemize}
    \item \textbf{1.1.1 General Web Search:} Broad internet search engines (Google, Bing, DuckDuckGo)
    \item \textbf{1.1.2 Academic and Research:} Scholarly databases, research repositories
    \item \textbf{1.1.3 News and Media:} Real-time news, media monitoring
    \item \textbf{1.1.4 Social Media Search:} Platform-specific content discovery
    \item \textbf{1.1.5 Specialized Directories:} Professional networks, business directories
\end{itemize}

\subsubsection*{1.2 Database and Repository Access}
\begin{itemize}
    \item \textbf{1.2.1 Structured Databases:} SQL databases, data warehouses
    \item \textbf{1.2.2 Document Repositories:} File systems, document management systems
    \item \textbf{1.2.3 Version Control:} Git repositories, code hosting platforms
    \item \textbf{1.2.4 Cloud Storage:} Object storage, file sharing services
    \item \textbf{1.2.5 External API Data Access:} REST API Clients, GraphQL Clients, Service-Specific APIs
\end{itemize}

\subsection*{2. MEMORY AND KNOWLEDGE MANAGEMENT}
\textbf{Definition:} Servers that store, organize, retrieve, and manage information for persistence and reuse.

\subsubsection*{2.1 Personal Knowledge Systems}
\begin{itemize}
    \item \textbf{2.1.1 Note-Taking Platforms:} Notion, Obsidian, Roam Research
    \item \textbf{2.1.2 Personal Wikis:} Individual knowledge bases
    \item \textbf{2.1.3 Bookmark Management:} Link organization and retrieval
\end{itemize}

\subsubsection*{2.2 Organizational Knowledge}
\begin{itemize}
    \item \textbf{2.2.1 Enterprise Knowledge Bases:} Corporate wikis, documentation systems
    \item \textbf{2.2.2 Collaborative Platforms:} Shared workspaces, team knowledge
    \item \textbf{2.2.3 Documentation Systems:} Technical documentation, help systems
\end{itemize}

\subsubsection*{2.3 Memory Persistence}
\begin{itemize}
    \item \textbf{2.3.1 Conversation Memory:} Chat history, interaction logs
    \item \textbf{2.3.2 Context Preservation:} Session state, user preferences
    \item \textbf{2.3.3 Learning Systems:} Adaptive knowledge accumulation
\end{itemize}

\subsubsection*{2.4 Knowledge Graphs and Ontologies}
\begin{itemize}
    \item \textbf{2.4.1 Semantic Networks:} RDF, OWL-based systems
    \item \textbf{2.4.2 Entity Relationship Systems:} Knowledge graph databases
    \item \textbf{2.4.3 Taxonomy Management:} Classification systems, controlled vocabularies
\end{itemize}

\subsection*{3. SIMULATION AND PLANNING}
\textbf{Definition:} Servers that model scenarios, predict outcomes, and generate strategic plans.

\subsubsection*{3.1 Computational Simulation}
\begin{itemize}
    \item \textbf{3.1.1 Mathematical Modeling:} Numerical simulations, equation solving
    \item \textbf{3.1.2 Physical Simulations:} Physics engines, material modeling
    \item \textbf{3.1.3 System Dynamics:} Complex system behavior modeling
\end{itemize}

\subsubsection*{3.2 Strategic Planning}
\begin{itemize}
    \item \textbf{3.2.1 Project Planning:} Task scheduling, resource allocation
    \item \textbf{3.2.2 Decision Support:} Multi-criteria analysis, option evaluation
    \item \textbf{3.2.3 Scenario Analysis:} What-if modeling, risk assessment
\end{itemize}

\subsubsection*{3.3 Predictive Analytics}
\begin{itemize}
    \item \textbf{3.3.1 Forecasting:} Time series prediction, trend analysis
    \item \textbf{3.3.2 Machine Learning Models:} Trained model inference
    \item \textbf{3.3.3 Statistical Analysis:} Hypothesis testing, correlation analysis
\end{itemize}

\subsection*{4. NAVIGATION AND MAPPING}
\textbf{Definition:} Servers that provide spatial awareness, location services, and navigation capabilities.

\subsubsection*{4.1 Geographic Information Systems}
\begin{itemize}
    \item \textbf{4.1.1 Mapping Services:} Google Maps, OpenStreetMap integrations
    \item \textbf{4.1.2 Geospatial Analysis:} GIS operations, spatial queries
    \item \textbf{4.1.3 Location Intelligence:} Place recognition, geocoding
\end{itemize}

\subsubsection*{4.2 Navigation and Routing}
\begin{itemize}
    \item \textbf{4.2.1 Route Planning:} Optimal path calculation
    \item \textbf{4.2.2 Real-time Navigation:} Turn-by-turn directions
    \item \textbf{4.2.3 Traffic and Conditions:} Dynamic routing with live data
\end{itemize}

\subsubsection*{4.3 Virtual Space Navigation}
\begin{itemize}
    \item \textbf{4.3.1 Digital Environment Mapping:} Virtual world navigation
    \item \textbf{4.3.2 Information Architecture:} Website/app structure navigation
    \item \textbf{4.3.3 Network Topology:} System architecture exploration
\end{itemize}

\subsection*{5. DATA EXTRACTION AND MANIPULATION}
\textbf{Definition:} Servers that process, transform, analyze, and manipulate various data types.

\subsubsection*{5.1 Text Processing}
\begin{itemize}
    \item \textbf{5.1.1 Natural Language Processing:} Parsing, analysis, generation
    \item \textbf{5.1.2 Document Processing:} OCR, format conversion, extraction
    \item \textbf{5.1.3 Content Analysis:} Sentiment, classification, summarization
\end{itemize}

\subsubsection*{5.2 Structured Data Operations}
\begin{itemize}
    \item \textbf{5.2.1 Data Transformation:} ETL processes, format conversion
    \item \textbf{5.2.2 Analytics and Reporting:} Statistical analysis, visualization
    \item \textbf{5.2.3 Database Operations:} CRUD operations, query execution
\end{itemize}

\subsubsection*{5.3 Multimedia Processing}
\begin{itemize}
    \item \textbf{5.3.1 Image Processing:} Computer vision, image manipulation, image generation
    \item \textbf{5.3.2 Audio Processing:} Speech recognition, speech to text, audio analysis
    \item \textbf{5.3.3 Video Processing:} Video analysis, content extraction
\end{itemize}

\subsubsection*{5.4 Web Data Extraction}
\begin{itemize}
    \item \textbf{5.4.1 Web Scraping:} HTML parsing, content extraction
    \item \textbf{5.4.2 API Data Harvesting:} Automated data collection
    \item \textbf{5.4.3 Real-time Monitoring:} Change detection, alert systems
\end{itemize}

\subsection*{6. REMOTE DEVICE CONTROL}
\textbf{Definition:} Servers that interface with and control external devices, systems, or IoT endpoints.

\subsubsection*{6.1 Smart Home and IoT}
\begin{itemize}
    \item \textbf{6.1.1 Home Automation:} Smart devices, environmental control
    \item \textbf{6.1.2 Security Systems:} Cameras, alarms, access control
    \item \textbf{6.1.3 Energy Management:} Smart meters, efficiency optimization
\end{itemize}

\subsubsection*{6.2 Industrial and Enterprise Systems}
\begin{itemize}
    \item \textbf{6.2.1 Network Infrastructure:} Router, switch, firewall management
    \item \textbf{6.2.2 Server Administration:} Remote system management
    \item \textbf{6.2.3 Manufacturing Control:} Industrial IoT, process control
\end{itemize}

\subsubsection*{6.3 Mobile and Wearable Devices}
\begin{itemize}
    \item \textbf{6.3.1 Smartphone Integration:} Device control, sensor access
    \item \textbf{6.3.2 Wearable Technology:} Fitness trackers, smartwatches
    \item \textbf{6.3.3 Location-based Services:} GPS, proximity systems
\end{itemize}

\subsection*{7. COMMUNICATION AND INTERACTION}
\textbf{Definition:} Servers that facilitate communication between entities, manage interactions, and handle messaging.

\subsubsection*{7.1 Messaging and Chat}
\begin{itemize}
    \item \textbf{7.1.1 Instant Messaging:} Real-time chat platforms
    \item \textbf{7.1.2 Email Integration:} Email sending, management, processing
    \item \textbf{7.1.3 Social Media Interaction:} Platform posting, engagement
\end{itemize}

\subsubsection*{7.2 Collaboration Tools}
\begin{itemize}
    \item \textbf{7.2.1 Video Conferencing:} Meeting platforms, screen sharing
    \item \textbf{7.2.2 Shared Workspaces:} Collaborative editing, project management
    \item \textbf{7.2.3 Workflow Automation:} Process orchestration, approval systems
\end{itemize}

\subsubsection*{7.3 Notification and Alerting}
\begin{itemize}
    \item \textbf{7.3.1 Push Notifications:} Mobile, desktop alerts
    \item \textbf{7.3.2 Monitoring Alerts:} System health, threshold notifications
    \item \textbf{7.3.3 Event Broadcasting:} Webhook delivery, event streaming
\end{itemize}

\subsection*{8. MULTI-DOMAIN ORCHESTRATION}
\textbf{Definition:} Servers that aggregate multiple tools, coordinate complex workflows, or provide meta-functionality across domains.

\subsubsection*{8.1 Tool Aggregators}
\begin{itemize}
    \item \textbf{8.1.1 Multi-Service Hubs:} Zapier-like integrations
    \item \textbf{8.1.2 Workflow Orchestrators:} Complex process automation
    \item \textbf{8.1.3 API Gateways:} Service mesh, API management
\end{itemize}

\subsubsection*{8.2 Meta-Programming Interfaces}
\begin{itemize}
    \item \textbf{8.2.1 Code Generation:} Automated programming assistance
    \item \textbf{8.2.2 System Integration:} Cross-platform connectivity
    \item \textbf{8.2.3 Configuration Management:} System setup, deployment
\end{itemize}

\subsubsection*{8.3 Hybrid Functionality Servers}
\begin{itemize}
    \item \textbf{8.3.1 Platform-Specific Suites:} Single-provider multi-tool servers
    \item \textbf{8.3.2 Domain-Crossing Tools:} Servers spanning multiple primary categories
    \item \textbf{8.3.3 Adaptive Interfaces:} Context-sensitive tool selection
\end{itemize}

\section*{Taxonomy-v2}

\subsection*{Approach}
\begin{itemize}
    \item \textbf{Primary Axis:} Functionality (What the server does)
    \item \textbf{Secondary Dimensions:} Data Type, Provider
\end{itemize}

\subsection*{Handling Multi-Tool Servers}
\begin{itemize}
    \item Allow servers to be tagged with multiple categories -- \textbf{PREFERRED}.
    \item This deviates from ``Single Inheritance'' but enables recognition of ambiguity and allows proposing alternatives or refining requests.
\end{itemize}

\subsection*{Primary Classification Rules}
\begin{itemize}
    \item \textbf{Dominant Function:} Each MCP server must have a \textbf{single, primary classification} based on its most prominent functionality (Level 1, 2, or 3).
    \item \textbf{Secondary Function:} Allow tagging the server with one or more additional functionalities.
    \item \textbf{Example:} A server with primary function ``1.1.1 General Web Search'' and secondary capability ``5.1.1 Natural Language Processing'' should be classified under 1.1.1 and tagged with 5.1.1.
    \item \textbf{``Other'' Categories:} Use \texttt{x.9} or \texttt{x.x.9} for broader uncategorized items.
\end{itemize}

\subsection*{Secondary Dimensions (Tags)}
These may be considered Level 4, but are treated as separate filters:
\begin{itemize}
    \item \textbf{Data Type:} Text, Image, Audio, Video, Multimodal, Structured, Unstructured
    \item \textbf{Provider:} Google, Microsoft, OpenAI, Custom, Open Source
\end{itemize}

\newpage

\subsection*{1. SEARCH AND INFORMATION RETRIEVAL}
\textbf{Definition:} Servers that locate, fetch, and deliver information from external sources including web search, databases, and external APIs.

\subsubsection*{1.1 Web Search and Discovery}
\begin{itemize}
    \item \textbf{1.1.1 General Web Search:} Broad internet search engines providing web results (e.g., Google, Bing, DuckDuckGo, general search APIs)
    \item \textbf{1.1.2 Academic and Research:} Specialized search for scholarly content including research papers, academic databases, scientific repositories, and educational resources
    \item \textbf{1.1.3 News and Media:} Real-time news aggregation, media monitoring, journalism databases, and current events tracking systems
    \item \textbf{1.1.4 Social Media Discovery:} Platform-specific content discovery across social networks
    \item \textbf{1.1.5 Specialized Directories:} Professional networks, business directories, industry-specific databases, and niche community platforms
\end{itemize}

\subsubsection*{1.2 Database and Repository Access}
\begin{itemize}
    \item \textbf{1.2.1 Structured Databases:} Direct access to SQL databases, data warehouses, and structured data stores
    \item \textbf{1.2.2 Document Repositories:} File systems, document management systems, enterprise content management, and document libraries
    \item \textbf{1.2.3 Version Control:} Git repositories, code hosting platforms, source control systems, and development collaboration tools
    \item \textbf{1.2.4 Cloud Storage:} Object storage services, file sharing platforms, cloud-based storage systems, and distributed file systems
    \item \textbf{1.2.5 External API Data Access:} REST API clients, GraphQL clients, service-specific APIs, and third-party integration endpoints
\end{itemize}

\subsection*{2. MEMORY AND KNOWLEDGE MANAGEMENT}
\textbf{Definition:} Servers that store, organize, retrieve, and manage internal information for persistence, reuse, and knowledge building.

\subsubsection*{2.1 Personal Knowledge Systems}
\begin{itemize}
    \item \textbf{2.1.1 Note-Taking Platforms:} Digital note-taking systems including Notion, Obsidian, Roam Research, and personal note management
    \item \textbf{2.1.2 Bookmark and Reference Management:} Link organization, reference collection, citation management, and personal library systems
\end{itemize}

\subsubsection*{2.2 Organizational Knowledge Systems}
\begin{itemize}
    \item \textbf{2.2.1 Enterprise Knowledge Bases:} Corporate wikis, institutional documentation systems, and organizational memory platforms
    \item \textbf{2.2.2 Collaborative Knowledge Platforms:} Shared workspaces, team knowledge systems, and collaborative documentation tools
    \item \textbf{2.2.3 Documentation and Help Systems:} Technical documentation, user guides, help desk knowledge bases, and support systems
\end{itemize}

\subsubsection*{2.3 Memory Persistence}
\begin{itemize}
    \item \textbf{2.3.1 Conversation Memory:} Chat history, interaction logs, session continuity, and dialogue state management
    \item \textbf{2.3.2 Context Preservation:} User preferences, personalization data, and settings management
    \item \textbf{2.3.3 Learning and Adaptation Systems:} Knowledge accumulation, user behavior learning, and adaptive interfaces
\end{itemize}

\subsubsection*{2.4 Knowledge Graphs and Semantic Systems}
\begin{itemize}
    \item \textbf{2.4.1 Semantic Networks:} RDF systems, OWL-based ontologies, linked data platforms, and semantic web technologies
    \item \textbf{2.4.2 Entity and Relationship Systems:} Knowledge graph databases, entity recognition, relationship mapping, and graph-based knowledge systems
\end{itemize}

\newpage

\subsection*{3. SIMULATION AND PLANNING}
\textbf{Definition:} Servers that model scenarios, predict outcomes, generate strategic plans, and support decision-making processes.

\subsubsection*{3.1 Computational Simulation}
\begin{itemize}
    \item \textbf{3.1.1 Mathematical Modeling:} Numerical simulations, equation solving, computational mathematics, and algorithmic modeling
    \item \textbf{3.1.2 Physical and Scientific Simulations:} Physics engines, material modeling, scientific simulations, and engineering analysis
    \item \textbf{3.1.3 System Dynamics:} Complex system behavior modeling, network simulations, and dynamic system analysis
\end{itemize}

\subsubsection*{3.2 Strategic Planning and Management}
\begin{itemize}
    \item \textbf{3.2.1 Project and Task Management:} Task scheduling, resource allocation, project planning tools, and workflow management
    \item \textbf{3.2.2 Decision Support Systems:} Multi-criteria analysis, option evaluation, strategic planning, and business intelligence
    \item \textbf{3.2.3 Scenario and Risk Analysis:} What-if modeling, risk assessment, contingency planning, and strategic scenario evaluation
\end{itemize}

\subsubsection*{3.3 Predictive Analytics}
\begin{itemize}
    \item \textbf{3.3.1 Forecasting and Trend Analysis:} Time series prediction, market forecasting, trend identification, and predictive modeling
    \item \textbf{3.3.2 Machine Learning Inference:} Trained model deployment, AI-powered predictions, and intelligent decision support
    \item \textbf{3.3.3 Statistical Analysis:} Hypothesis testing, correlation analysis, data mining, and statistical modeling
\end{itemize}

\subsection*{4. NAVIGATION AND MAPPING}
\textbf{Definition:} Servers that provide spatial awareness, location services, navigation capabilities, and geographic information systems.

\subsubsection*{4.1 Geographic Information Systems}
\begin{itemize}
    \item \textbf{4.1.1 Mapping Services:} Google Maps, OpenStreetMap integrations, cartographic services, and geographic visualization
    \item \textbf{4.1.2 Geospatial Analysis:} GIS operations, spatial queries, geographic data processing, and location intelligence
    \item \textbf{4.1.3 Location and Place Services:} Geocoding, place recognition, address validation, and location-based information
\end{itemize}

\subsubsection*{4.2 Physical Navigation and Routing}
\begin{itemize}
    \item \textbf{4.2.1 Route Planning:} Optimal path calculation, multi-modal routing, and travel planning systems
    \item \textbf{4.2.2 Real-time Navigation:} Turn-by-turn directions, live traffic integration, and dynamic route guidance
    \item \textbf{4.2.3 Traffic and Environmental Conditions:} Real-time traffic data, weather-aware routing, and condition-based navigation
\end{itemize}

\subsubsection*{4.3 Virtual and Digital Navigation}
\begin{itemize}
    \item \textbf{4.3.1 Digital Environment Mapping:} Virtual world navigation, game environment mapping, and 3D space orientation
    \item \textbf{4.3.2 Information Architecture Navigation:} Website structure exploration, app navigation assistance, and content hierarchy mapping
    \item \textbf{4.3.3 Network and System Topology:} Infrastructure mapping, system architecture exploration, and network navigation
\end{itemize}

\subsection*{5. DATA EXTRACTION AND MANIPULATION}
\textbf{Definition:} Servers that process, transform, analyze, and manipulate various data types including text, multimedia, and structured data.

\subsubsection*{5.1 Text and Document Processing}
\begin{itemize}
    \item \textbf{5.1.1 Natural Language Processing:} Text parsing, language analysis, generation, translation, and linguistic operations
    \item \textbf{5.1.2 Document Processing:} OCR, format conversion, document extraction, and file format manipulation
    \item \textbf{5.1.3 Content Analysis:} Sentiment analysis, text classification, summarization, and content understanding
\end{itemize}

\subsubsection*{5.2 Structured Data Operations}
\begin{itemize}
    \item \textbf{5.2.1 Data Transformation:} ETL processes, format conversion, data cleaning, and structural data manipulation
    \item \textbf{5.2.2 Analytics and Reporting:} Statistical analysis, data visualization, business intelligence, and reporting systems
    \item \textbf{5.2.3 Database Operations:} CRUD operations, query execution, database management, and data persistence
\end{itemize}

\subsubsection*{5.3 Multimedia Processing}
\begin{itemize}
    \item \textbf{5.3.1 Image Processing:} Computer vision, image manipulation, image generation, and visual content analysis
    \item \textbf{5.3.2 Audio Processing:} Speech recognition, audio analysis, sound processing, and voice-to-text conversion
    \item \textbf{5.3.3 Video Processing:} Video analysis, content extraction, video editing, and multimedia content processing
\end{itemize}

\subsubsection*{5.4 Web Data Extraction}
\begin{itemize}
    \item \textbf{5.4.1 Web Scraping:} HTML parsing, content extraction, web crawling, and automated data harvesting
    \item \textbf{5.4.2 API Data Collection:} Automated data gathering from APIs, bulk data retrieval, and systematic data acquisition
    \item \textbf{5.4.3 Monitoring and Change Detection:} Real-time monitoring, alert systems, and automated change tracking
\end{itemize}

\subsection*{6. SYSTEM AND DEVICE CONTROL}
\textbf{Definition:} Servers that interface with and control external devices, systems, applications, or IoT endpoints.

\subsubsection*{6.1 Smart Home and IoT}
\begin{itemize}
    \item \textbf{6.1.1 Home Automation:} Smart devices, environmental control, lighting systems, and household device management
    \item \textbf{6.1.2 Security and Monitoring:} Cameras, alarms, access control, and home security systems
    \item \textbf{6.1.3 Energy and Utilities Management:} Smart meters, efficiency optimization, utility monitoring, and resource management
\end{itemize}

\subsubsection*{6.2 Enterprise and Infrastructure Systems}
\begin{itemize}
    \item \textbf{6.2.1 Network Infrastructure:} Router, switch, firewall management, and network device configuration
    \item \textbf{6.2.2 Server and System Administration:} Remote system management, server operations, and infrastructure control
    \item \textbf{6.2.3 Industrial and Manufacturing Control:} Industrial IoT, process control, manufacturing systems, and operational technology
\end{itemize}

\subsubsection*{6.3 Computer and Application Control}
\begin{itemize}
    \item \textbf{6.3.1 Operating System Control:} Desktop automation, system-level operations, OS-specific integrations (Windows, macOS, Linux)
    \item \textbf{6.3.2 Browser and Application Control:} Web browser automation, application scripting, and software control interfaces
    \item \textbf{6.3.3 Mobile and Wearable Devices:} Smartphone integration, device sensor access, wearable technology, and mobile device management
\end{itemize}

\subsection*{7. COMMUNICATION AND INTERACTION}
\textbf{Definition:} Servers that facilitate communication between entities, manage interactions, handle messaging, and coordinate collaborative activities.

\subsubsection*{7.1 Messaging and Social Interaction}
\begin{itemize}
    \item \textbf{7.1.1 Instant Messaging:} Real-time chat platforms, messaging apps, and communication channels
    \item \textbf{7.1.2 Email Integration:} Email sending, management, processing, and email automation systems
    \item \textbf{7.1.3 Social Media Interaction:} Platform posting, social engagement, community management, and social media automation
\end{itemize}

\subsubsection*{7.2 Collaboration and Scheduling}
\begin{itemize}
    \item \textbf{7.2.1 Video Conferencing:} Meeting platforms, screen sharing, virtual collaboration, and video communication tools
    \item \textbf{7.2.2 Calendar and Scheduling:} Meeting scheduling, calendar management, appointment booking, and time coordination
    \item \textbf{7.2.3 Shared Workspaces:} Collaborative editing, team coordination, document sharing, and group productivity tools
\end{itemize}

\subsubsection*{7.3 Workflow and Process Automation}
\begin{itemize}
    \item \textbf{7.3.1 Workflow Automation:} Process orchestration, approval systems, business process automation, and workflow management
    \item \textbf{7.3.2 Event-based Alerts:} Push notifications, system alerts, event broadcasting, and notification management
    \item \textbf{7.3.3 Event and Integration Management:} Webhook delivery, event streaming, system integration, and inter-service communication
\end{itemize}

\subsection*{8. SPECIALIZED DOMAINS}
\textbf{Definition:} Domain-specific servers tailored for particular industries, activities, or specialized use cases.

\subsubsection*{8.1 Financial Services}
\begin{itemize}
    \item \textbf{8.1.1 Personal and Business Finance:} Budget planning, expense tracking, financial health analysis, accounting integration
    \item \textbf{8.1.2 Trading and Investments:} Stock trading, crypto, portfolio management, and market research
    \item \textbf{8.1.3 Payment Processing:} Payment gateways, money transfers, digital wallets, and payment services
    \item \textbf{8.1.4 Market Data and Analytics:} Financial data feeds, economic indicators, and financial research
\end{itemize}

\subsubsection*{8.2 Entertainment and Gaming}
\begin{itemize}
    \item \textbf{8.2.1 Gaming Platforms:} Game APIs, player statistics, platform integrations
    \item \textbf{8.2.2 Game Development:} Game engines, development frameworks, and tools
    \item \textbf{8.2.3 Media and Entertainment:} Streaming platforms, music, podcasts, and content recommendation
\end{itemize}

\subsubsection*{8.9 Other Specialized Domains}
\begin{itemize}
    \item \textbf{Definition:} Domains like Shopping, Food, Health, or Travel not covered above.
\end{itemize}

\subsection*{9. DEVELOPER TOOLS AND PROGRAMMING}
\textbf{Definition:} Servers that assist software development, provide programming utilities, code management, and development workflow support.

\subsubsection*{9.1 Code Development and Management}
\begin{itemize}
    \item \textbf{9.1.1 Code Generation and AI Assistance:} Automated programming, code completion
    \item \textbf{9.1.2 Code Analysis and Quality:} Review, testing frameworks, quality tools
    \item \textbf{9.1.3 Development Environment:} IDE integrations, toolchains
    \item \textbf{9.1.4 API Development and Testing:} API design, documentation, and testing
    \item \textbf{9.1.5 Development Utilities:} Build tools, package and dependency management
\end{itemize}

\subsubsection*{9.2 Development Operations}
\begin{itemize}
    \item \textbf{9.2.1 Deployment and CI/CD:} Integration pipelines, deployment automation
    \item \textbf{9.2.2 Infrastructure as Code:} Provisioning, configuration automation
    \item \textbf{9.2.3 Monitoring and Debugging:} Application monitoring, performance analysis
    \item \textbf{9.2.4 Integration and Connectivity:} Middleware, service mesh, integration platforms
\end{itemize}

\subsection*{10. MULTI-DOMAIN ORCHESTRATION}
\textbf{Definition:} Servers that aggregate multiple tools, coordinate complex workflows across domains, or provide meta-functionality spanning multiple categories.

\subsubsection*{10.1 Integration and Orchestration}
\begin{itemize}
    \item \textbf{10.1.1 Multi-Service Integration:} Zapier-like tools, API gateways, cross-platform workflows
    \item \textbf{10.1.2 Workflow and Process Orchestration:} Enterprise workflow systems, BPM, approval workflows
\end{itemize}

\subsubsection*{10.2 Cross-Domain and Platform Tools}
\begin{itemize}
    \item \textbf{10.2.1 Platform-Specific Suites:} Vendor ecosystems, integrated toolsets
    \item \textbf{10.2.2 Domain-Crossing Applications:} Servers spanning multiple primary categories
\end{itemize}

\subsection*{11. Other}
\textbf{Definition:} Servers that do not fall into any of the available categories.

\end{document}